\documentstyle[colap]{article}

\def\drafter{{\sc drafter}}
\begin{document}

\newenvironment{bullist}
    {\begin{list}{$\bullet$}
	{\parsep 0pt \itemsep 0pt \setlength{\rightmargin}{\leftmargin}}}%
    {\end{list}}

\newenvironment{examples}{\par\begin{enumerate}}{\end{enumerate}}

\input{psfig}

\title{\submitted{INLG96 ----  {\em 1996 International Workshop on Natural
Language Generation}, Herstmonceux, England}
Learning Micro-Planning Rules for Preventative
Expressions\thanks{This work is partially supported by the Engineering
and Physical Sciences Research Council ({\sc epsrc}) Grant
\mbox{J19221}, by {\sc bc/daad arc} Project 293, and by the Commission
of the European Union Grant \mbox{{\sc lre}-62009}.}}
\author{\parbox[t]{3in}{\centering{\bf Keith Vander
Linden}\thanks{After September 1, Dr. Vander Linden's
address will be Department of Mathematics and Computer Science, Calvin
College, Grand Rapids, MI 49546, USA.} \\
	Information Technology Research Institute\\
	University of Brighton\\
	Brighton BN2 4AT, UK\\
	{\em email:\/} knvl@itri.brighton.ac.uk}
	\parbox[t]{3in}{\centering{\bf Barbara Di Eugenio}\\
	Computational Linguistics\\
	Carnegie Mellon University\\
	Pittsburgh, PA, 15213  USA\\
	{\em email:\/} dieugeni@andrew.cmu.edu}}
\date{}

\makeatletter
 \def\submitted#1{\setbox\@tempboxa\vbox{\normalsize \tt \raggedright
    #1 \\ \hbox{}}
    \vspace{-1.5 cm} \usebox\@tempboxa \\
    \vspace{-\ht\@tempboxa} \vspace{1.5 cm}}
\makeatother

\maketitle
\thispagestyle{empty}

\section*{Abstract}

Building text planning resources by hand is time-consuming and
difficult.  Certainly, a number of planning architectures and their
accompanying plan libraries have been implemented, but while the
architectures themselves may be reused in a new domain, the library of
plans typically cannot.  One way to address this problem is to use
machine learning techniques to automate the derivation of planning
resources for new domains.  In this paper, we apply this technique to
build micro-planning rules for preventative expressions in
instructional text.

\section{Introduction}

Building text planning resources by hand is time-consuming and
difficult.  Certainly, much work has been done in this regard; there
are a number of freely available text planning architectures (e.g.,
\cite{moore-paris-cl93}).  It is frequently the case, however, that
while the architecture itself can be reused in a new domain, the
library of text plans developed for it cannot.  In particular,
micro-planning rules, those rules that specify the low-level
grammatical details of expression, are highly sensitive to variations
between sub-languages, and are therefore difficult to reuse.

When faced with a new domain in which to generate text, the typical
scenario is to perform a corpus analysis on a representative
collection of the text produced by human authors in that domain and to
induce a set of micro-planning rules guiding the generation process in
accordance with the results.  Some fairly simple rules usually jump
out of the analysis quickly, mostly based on the analyst's intuitions.
For example, in written instructions, user actions are typically
expressed as imperatives.  Such observations, however, tend to be
gross characterisations.  More accurate micro-planning requires
painstaking analysis.  In this paper, for example, the micro-planner
must distinguish between phrasing such as ``Don't do {\em action-X\/}''
and ``Take care not to do {\em action-X\/}''.  Without analysis, it is
far from clear how this decision can best be made.

Some form of automation would clearly be desirable.  Unfortunately,
corpus analysis techniques are not yet capable of automating the
initial phases of the corpus study (nor will they be for the
foreseeable future).  There are, however, techniques for rule
induction which are useful for the later stages of corpus analysis and
for implementation.

In this paper, we focus on the use of such rule induction techniques
in the context of the micro-planning of preventative expressions in
instructional text.  We define what we mean by a preventative
expression, and go on to describe a corpus analysis in which we derive
three features that predict the grammatical form of such expressions.
We then use the C4.5 learning algorithm to construct a micro-planning
sub-network appropriate for these expressions.  We conclude with an
implemented example in which the technical author is allowed to set
the relevant features, and the system generates the appropriate
expressions in English and in French.

\section{Preventative Expressions}

Preventative expressions are used to warn the reader not to perform
certain inappropriate or potentially dangerous actions.  The reader
may be told, for example, ``Do not enter'' or ``Take care not to push
too hard''.  Both of these examples involve negation (``do {\em
not\/}'' and ``take care {\em not\/}'').  Although this is not
strictly necessary for preventative expressions (e.g., one might say
``stay out'' rather than ``do not enter''), we will focus on the use
of negative forms in this paper, using the following
categorisation:\footnote{Horn (\shortcite{horn89:linguistics}) gives a
more complete categorisation of negative forms.}

\begin{itemize}
\item  negative imperatives proper (termed {\em DONT\/} imperatives)
--- These are characterised by the negative auxiliary {\em do not\/} or
	{\em don't\/}, as in:
  \begin{examples} 
  \item Your sheet vinyl floor may be vinyl asbestos,
  which is no longer on the market. {\em Don't sand it or tear
  it up\/} because this will put dangerous asbestos fibers into the air.
  \end{examples}
\item  {\em NEVER\/} imperatives  --- These are characterised by the use
of the negative adverb {\em never\/}, as in:
  \begin{examples} 
  \item Whatever you do, {\em never go to Vienna if you are on a diet.\/}
  \end{examples}
\item other negative imperatives (termed {\em neg-TC\/}
imperatives) --- These include {\em take care\/} and {\em be careful\/}
followed by a negative infinitival complement, as in the following
examples:
  \begin{examples}
  \item To book the strip, fold the bottom third or more of the strip 
  over the middle of the panel, pasted sides together, {\em taking care
  not to crease the wallpaper sharply at the fold.}
  \item If your plans call for replacing the wood base molding with vinyl
  cove molding, {\em be careful not to damage the walls\/} as you remove
  the wood base.
  \end{examples}
\end{itemize}

\section{Corpus Analysis}

In terms of text generation, our interest is in finding mappings from
features related to the {\em function\/} of these expressions, to
those related to their grammatical {\em form\/}.  Functional features
include the semantic features of the message being expressed, the
pragmatic features of the context of communication, and the features
of the surrounding text being generated.  In this section we will
briefly discuss the nature of our corpus, and the function and form
features that we have coded.  We will conclude with a discussion of
the inter-coder reliability.  A more detailed discussion of this
portion of the work is given elsewhere (\cite{preventions-coling96}).

\subsection{Corpus}\label{corpus}

The corpus from which we take all our coded examples has been
collected opportunistically off the internet and from other sources.
It is 4.5 MB in size and is made entirely of written English
instructional texts. As a collection, these texts are the result of a
variety of authors working in a variety of contexts.

We broke the corpus texts into expressions using a simple sentence
breaking algorithm and then collected the negative imperatives by
probing for expressions that contain the grammatical forms we were
interested in (i.e., expressions containing phrases such as {\em
don't}, {\em never}, and {\em take care\/}).  The grammatical forms we
found, 1283 occurrences in all, constitute 2.7\% of the expressions in
the full corpus.  The first line in Table~\ref{neg-table}, marked
``Raw Grep'', indicates the quantity of each type.  

\begin{table*}[t]
\centering
\begin{tabular}{||l||c|c||c||c|c|c|c||}\hline 
& \multicolumn{2}{c||}{DONT} & {NEVER} & \multicolumn{4}{c||}{Neg-TC}  \\ \hline
& {\em don't\/} & {\em do not\/} &  & {\em take care\/} & {\em make sure\/} &
{\em be careful\/} & {\em be sure\/} \\ \hline \hline
Raw Grep & 417 & 385 & 108 & 21 & 229 & 52 & 71 \\ \hline 
Raw Sample & 100 & 99 & 108 & 21 & 104 & 52 & 71\\ \hline 
Final Coding & 78 & 89 & 40 & 17 & 3 & 46 & 6\\ \hline 
& \multicolumn{2}{c||}{167} & 40 & \multicolumn{4}{c||}{72}  \\ \hline
\end{tabular}
\caption{Distribution of negative imperatives}
\label{neg-table}
\end{table*}

We then filtered the results.  When the probe returned more than 100
examples for a grammatical form, we randomly selected around 100 of
those returned, as shown in line 2 of Table~\ref{neg-table} (labelled
``Raw Sample'').  We then removed those examples that, although they
contained the desired lexical string, did not constitute negative
imperatives (e.g., ``If you {\em don't\/} like the colors of the file,
\ldots, use Binder to change them.''), as shown in line 3, labelled
``Final Coding''.  

\begin{table*}[t]
\centering
\begin{tabular}{||l|c|c||}\hline 
{\em Instruction type\/} & {\em Corpus size\/} & {\em \# of preventatives\/} \\ \hline
Recipes & 1.7M & 83 \\ 
Do-it-yourself & 1.26M & 99 \\ 
Di Eugenio's thesis\footnotemark & 336K & 69 \\ 
Software instructions & 264K & 0 \\ 
Administrative forms & 317K & 9 \\ 
Other & 565K & 19 \\ \hline\hline
{\em Totals\/} & 4.5M & 279 \\ \hline
\end{tabular}
\caption{Distribution of examples from sample}
\label{distribution-table}
\end{table*}

The final corpus sample is made up of 279 examples, all of which have
been coded for the features to be discussed in the next two sections.
Table~\ref{distribution-table} also shows the relative sizes of the
various types of instructions in the corpus as well as the number of
examples from this sample that came from each type.

\subsection{Form}

\footnotetext{Note that we used a number of
examples from Di Eugenio's thesis (\shortcite{dieugenio-thesis}) which
were included as excerpts.  In this table we include only an estimate
of the full size of that portion of the corpus.}

Because of its syntactic nature, the form feature coding was very
robust.  The possible feature values were: {\bf DONT} --- for the {\em
do not\/} and {\em don't\/} forms discussed above; {\bf NEVER}, for
imperatives containing {\em never\/}; and {\bf neg-TC} --- for {\em
take care\/}, {\em make sure\/}, {\em be careful\/}, and {\em be
sure\/} expressions with negative arguments.  The two authors agreed
on their coding of this feature in all cases.

\subsection{Function Features}\label{function-features}

We will now briefly discuss three of the function features we have
coded: {\sc intentionality}, {\sc awareness}, and {\sc safety}. We
illustrate them in turn using $\alpha$ to refer to the prevented
action and using ``agent'' to refer to the reader and executer of the
instructions.

\paragraph{\bf Intentionality:}

This feature encodes whether or not the writer believes that the agent
will consciously adopt the intention of performing $\alpha$:

\begin{description}
\item[{\sc CON}] is used to code situations where 
the agent intends to perform $\alpha$.  In this case, the agent must
be aware that $\alpha$ is one of his or her possible alternatives.

\item[{\sc UNC}] is used to code situations in which 
the agent doesn't realize that there is a choice involved
(cf. \cite{dieugenio93:discourse}).  It is used in
two situations: when $\alpha$ is totally accidental, or
\end{description}

\paragraph{\bf Awareness:}

This feature captures whether or not the writer believes that the
agent is aware that the consequences of $\alpha$ are bad:

\begin{description}

\item[{\sc AW}] is used when the agent is aware that $\alpha$ is bad.
For example, the agent may be told ``Be careful not to burn the
garlic'' when he or she is perfectly well aware that burning things
when cooking them is bad.

\item[{\sc UNAW}] is used when the agent is perceived to be unaware 
that $\alpha$ is bad.  
\end{description}

\paragraph{\bf Safety:}

This feature captures whether or not the author believes that the
agent's safety is put at risk by performing $\alpha$:

\begin{description}

\item[{\sc BADP}] is used when the agent's safety is put at risk by
performing $\alpha$.

\item[{\sc NOT}] is used when it is not unsafe to perform  $\alpha$,
but may, rather, be simply inconvenient.

\end{description}

\subsection{Inter-coder reliability}\label{inter}

Each author independently coded each of the features for all the
examples in the sample.  The percentage agreement for each of the
features is shown in the following table:

\begin{center}
\begin{tabular}{||l|c||}\hline
{\em feature\/} & {\em percent agreement\/}  \\ \hline 
form & 100\% \\ 
\hline
intentionality & 74.9\% \\
awareness & 93.5\% \\ 
safety & 90.7\% \\ 
\hline
\end{tabular}
\end{center}

As advocated by Carletta (\shortcite{carletta96}), we have used the
Kappa coefficient (\cite{siegel88}) as a measure of coder agreement.
For nominal data, this statistic not only measures agreement, but also
factors out chance agreement.

If $P(A)$ is the proportion of times the coders agree, and $P(E)$ is
the proportion of times that coders are expected to agree by chance, K
is computed as follows:

$$ {K \: = \: {\frac{P(A)\,-\,P(E)}{1\,-\,P(E)}}} $$

There are various ways of computing $P(E)$ according to Siegel and
Castellan (\shortcite{siegel88}); most researchers agree on the
following formula, which we also adopted: $$P(E) \: = \:
{\sum_{j=1}^{m} p_{j}^{2}} $$ where $m$ is the number of categories,
and $p_{j}$ is the proportion of objects assigned to category $j$.

The mere fact that K may have a value $k$ greater than zero is not
sufficient to draw any conclusion, however, as it must be established
whether $k$ is significantly different from zero. There are
suggestions in the literature that allow us to draw general
conclusions without these further computations. For example, Rietveld
and van Hout (\shortcite{rietveld93}) suggest the correlation between
K values and inter-coder reliability shown in the following table:

\begin{center}
\begin{tabular}{|| l | l ||} \hline
{\em Kappa Value\/} & {\em Reliability Level\/} \\ \hline
.00 --  .20 & slight \\
.21 --  .40 & fair \\
.41 --  .60 & moderate \\
.61 --  .80 & substantial \\
.81 --	 1.00 & almost perfect \\ \hline
\end{tabular}
\end{center}

\noindent
For the form feature, the Kappa value is 1.0, indicating perfect
agreement.  The function features, which are more subjective in
nature, engender more disagreement among coders, as shown by the K
values in the following table:

\begin{center}
\begin{tabular}{||l|c||}\hline
{\em feature\/} & {\em K\/} \\ \hline 
{\sc intentionality} & 0.46 \\
{\sc awareness} & 0.76 \\ 
{\sc safety} & 0.71  \\ 
\hline
\end{tabular}
\end{center}

\noindent
According to this table, therefore, the {\sc awareness} and {\sc
safety} features show ``substantial'' agreement and the {\sc
intentionality} feature shows ``moderate'' agreement.  We have coded
other functional features as well, but they have either not proven as
reliable as these, or are not as useful in text planning.

In addition, Siegel and Castellan (\shortcite{siegel88}) point out
that it is possible to check the significance of K when the number of
objects is large; this involves computing the distribution of K
itself.  Under this approach, the three values above are significant
at the .000005 level.

\section{Automated Learning}

The corpus analysis results in a set of examples coded with the values
of the function and form features.  This data can be used to find
correlations between the two types of features, correlations, which,
in text generation, are typically implemented as decision trees or
rule sets mapping from function features to forms.

In this study, we used 179 coded examples as input to the learning
algorithm.  These are the examples on which the two authors agreed on
their coding of all the features.  The distribution of the grammatical
forms in these examples is shown in the following table:

\begin{center}
\begin{tabular}{||l|c||}\hline
{\em form\/} & {\em frequency\/} \\ \hline 
DONT & 100 \\
Neg-TC & 57 \\ 
NEVER & 22  \\ 
\hline
\end{tabular}
\end{center}

\noindent
The learning algorithm used these examples to derive a decision tree
which we then integrated into an existing micro-planner.

\subsection{Data Mining}

\begin{figure*}[t]
\begin{center}
\strut{\psfig{figure=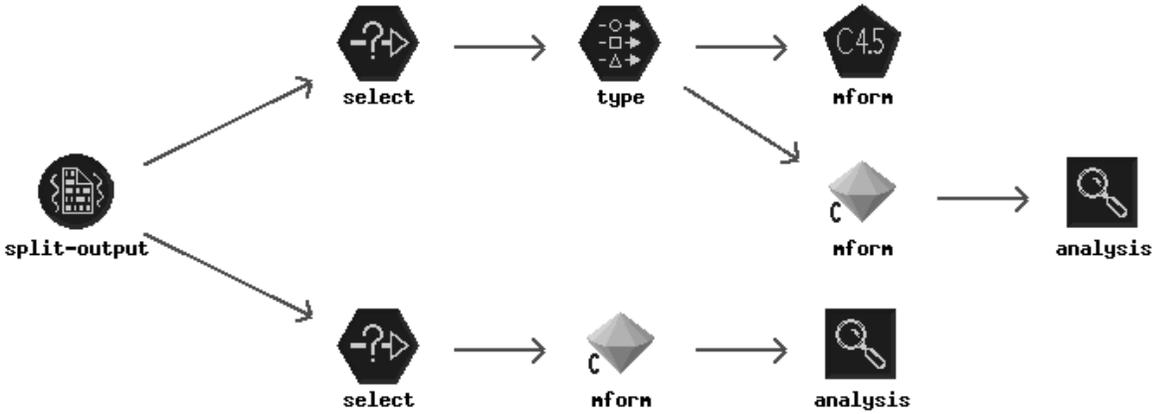,width=\textwidth}}
\caption{The Clementine learning environment}
\label{stream}
\end{center}
\end{figure*}

We have used Quinlan's C4.5 learning algorithm
(\shortcite{quinlan93:other}) in this study; this algorithm can induce
either decision trees or rules.  To provide a more convenient learning
environment, we have used Clementine (\shortcite{clementine}), a tool
which allows rapid reconfiguration of various data manipulation
facilities, including C4.5.  Figure~\ref{stream} shows the basic
control stream we used for learning and testing decision trees.  Data
is input from the {\bf split-output} file node on the left of the
figure and is passed through filtering modules until it reaches the
output modules on the right.  The two {\bf select} modules (pointed to
by the main input node) select the examples reserved for the training
set and the testing set respectively.  The upper stream processes the
training set and contains a {\bf type} module which marks the main
syntactic form (i.e., DONT, NEVER, or Neg-TC) as the variable to be
predicted and the {\sc awareness}, {\sc safety}, and {\sc
intentionality} features as the inputs.  Its output is passed to the
C4.5 node, labelled {\bf mform}, which produces the decision tree.  We
then use two copies of the resulting decision tree, represented by the
diamond shaped nodes marked with {\bf mform}, to test the accuracy of
the testing and the training sets.

One run of the system, for example, gave the following decision tree:

\begin{quote}
\begin{verbatim}
awareness = AW: NEG-TC 
awareness = UNAW:
|   intention = CON: DONT 
|   intention = UNC:
|   |   safety = BADP: NEVER 
|   |   safety = NOT: DONT 
\end{verbatim}
\end{quote}

\noindent
This tree takes the three function features and predicts the DONT,
NEVER, and Neg-TC forms.  It confirms our intuitions that {\em never\/}
imperatives are used when personal safety may be endangered (coded as
safety$=$``BADP''), and that Neg-TC forms are used when the reader is
expected to be aware of the danger that may arise
(cf. \cite{preventions-coling96}).  It accurately predicts the
grammatical form of 74.5\% of the 161 training examples, and 83.3\% of
the 18 testing examples.

Because there are relatively few training examples in our coded
corpus, we have also performed a 10-way cross-validation
test.\footnote{A cross-validation test is a test where C4.5 breaks the
data into different combinations of training and testing sets, builds
and tests decision trees for each, and averages the results
(\cite{clementine}).}  None of the derived trees in this test were
remarkably different from the one just shown, although they did order
the {\sc intentionality} and {\sc awareness} features differently.
The average accuracy of the learned decision trees on the testing sets
was 75.4\%.

Note that although this level of accuracy is better than 55.9\%, the
score achieved by simply selecting DONT in all cases, there is still
more work to be done.  The current features must be refined, and more
features may be need to be added.  We are currently experimenting with
a number of possibilities.  Note also that we have not distinguished
between the various sub-forms of DONT and Neg-TC shown in
Table~\ref{neg-table}; this will require yet more features.

Clementine can also ``balance'' the input to C4.5 by duplicating
training examples with under-represented feature values.  We used this
to increase the number of NEVER and Neg-TC examples to match the
number of DONT examples.  Ultimately, this reduced the accuracy of the
learned trees to 68.0\% in a cross-validation test.  The resulting
decision trees tended not to include all three features.

\subsection{Integration}

Because it is common for us to rebuild decision trees frequently
during analysis, we implemented a routine which automatically converts
the decision tree into the appropriate KPML-style system networks with
their associated choosers, inquiries, and inquiry implementations
(\cite{kpml-manual}).  This makes the network compatible with the
\drafter\ micro-planner, a descendent of {\sc imagene}
(\cite{vanderlinden-cl95}).  The conversion routine takes the
following inputs:

\begin{itemize}
\item the applicable language(s) --- C4.5 produces its decision trees
based on examples from a particular language, and KPML is capable of
being conditionalised for particular languages.  Thus, we may perform
separate corpus analyses of a particular phenomenon for various
languages, and learn separate micro-planning trees;
\item the input feature(s) --- The sub-network being built
must fit into the overall categorisations of the full micro-planner,
and thus we must specify the text functions that would trigger entry
to the new sub-network;
\item the decision tree itself;
\item a feature-value function  --- To traverse the new sub-network,
the KPML inquiries require a function that can determine the value of
the features for each pass through the network;
\item grammatical form specifications --- The sub-network must
eventually build sentence plan language (SPL) commands for input to
KPML, and thus must be told the appropriate SPL terms to use to
specify the required grammatical forms;
\item an output file name.
\end{itemize}

\noindent
For our example, the system sub-network shown in
Figure~\ref{system-network} is produced based on the decision tree
shown above.\footnote{Only the systems are shown in the KPML dump
given in Figure~\ref{system-network}.  The realisation statements,
choosers, inquiries, and inquiry implementations are not shown.}  It
is important to note here that although the micro-planner is
implemented as a systemic resource, the machine learning algorithm is
no respecter of systemic linguistic theory.  It simply builds decision
trees.  This gives rise to three distinctly non-systemic features of
these learned networks:

\begin{figure*}[t]
\begin{center}
\strut{\psfig{figure=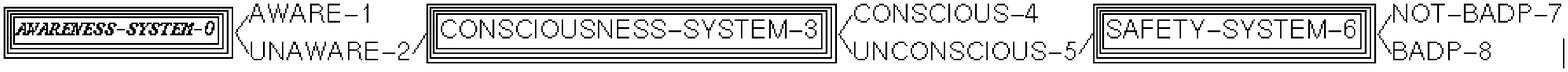,width=\textwidth}}
\caption{The micro-planner system network derived from the decision tree}
\label{system-network}
\end{center}
\end{figure*}

\begin{enumerate}
\item The realisation statements are included only at the leaf nodes
of the network.  We have built no intelligent facility for decomposing
the realisation statements and filtering common realisations up the
tree.
\item The learning algorithm will freely reuse systems (i.e.,
features) as various points in the tree.  This did not happen in
Figure~\ref{system-network}, but occasionally one of the features is
independently used in different sub-trees of the network.  We are
forced, therefore, to index the system and feature names with integers
to disambiguate.
\item There is no meta-functional distinction in the network, but
rather, all the features, regardless of their semantic type, are
included in the same tree.
\end{enumerate}

The sub-network derived in this section was spliced into the existing
micro-planning network for the full generation system.  As mentioned
above, this integration was done by manually specifying the desired
input conditions for the sub-network when the micro-planning rules are
built.  For the preventative expression sub-network, this turned out
to be a relatively simple matter.  \drafter's model of procedural
relations includes a {\em warning\/} relation which may be attached by
the author where appropriate.  The micro-planner, therefore, is able
to identify those portions of the procedure which are to be expressed
as warnings, and to enter the derived sub-network appropriately.  This
same process could be done with any of the other procedural relations
(e.g., purpose, precondition).  This assumes, however, the existence
of a core set of micro-plans which perform the procedural
categorisation properly; these were built by hand.  We have only just
begun to experiment with the possibility of building the entire
network automatically from a more exhaustive corpus analysis.

\section{A \drafter\ Example}

Given the corpus analysis and the learned system networks discussed
above, we will present an example of how preventative expressions can
be delivered in \drafter, an implemented text generation application.
\drafter\ is a instructional text authoring tool that allows
technical authors to specify a procedural structure, and then uses
that structure as input to a multilingual text generation facility
(\cite{drafter-ieee96}).  The instructions are generated in English
and in French.

To date, our domain of application has been manuals for software user
interfaces, but because this domain does not commonly contain
preventative expressions (see Table~\ref{distribution-table}), we
have extended \drafter's domain model to include coverage for
do-it-yourself applications.  Although this switch has entailed some
additions to the domain model, \drafter's input and generation
facilities remain as they were.

\subsection{Input Specification}

\begin{figure*}[t]
\begin{center}
\strut{\psfig{figure=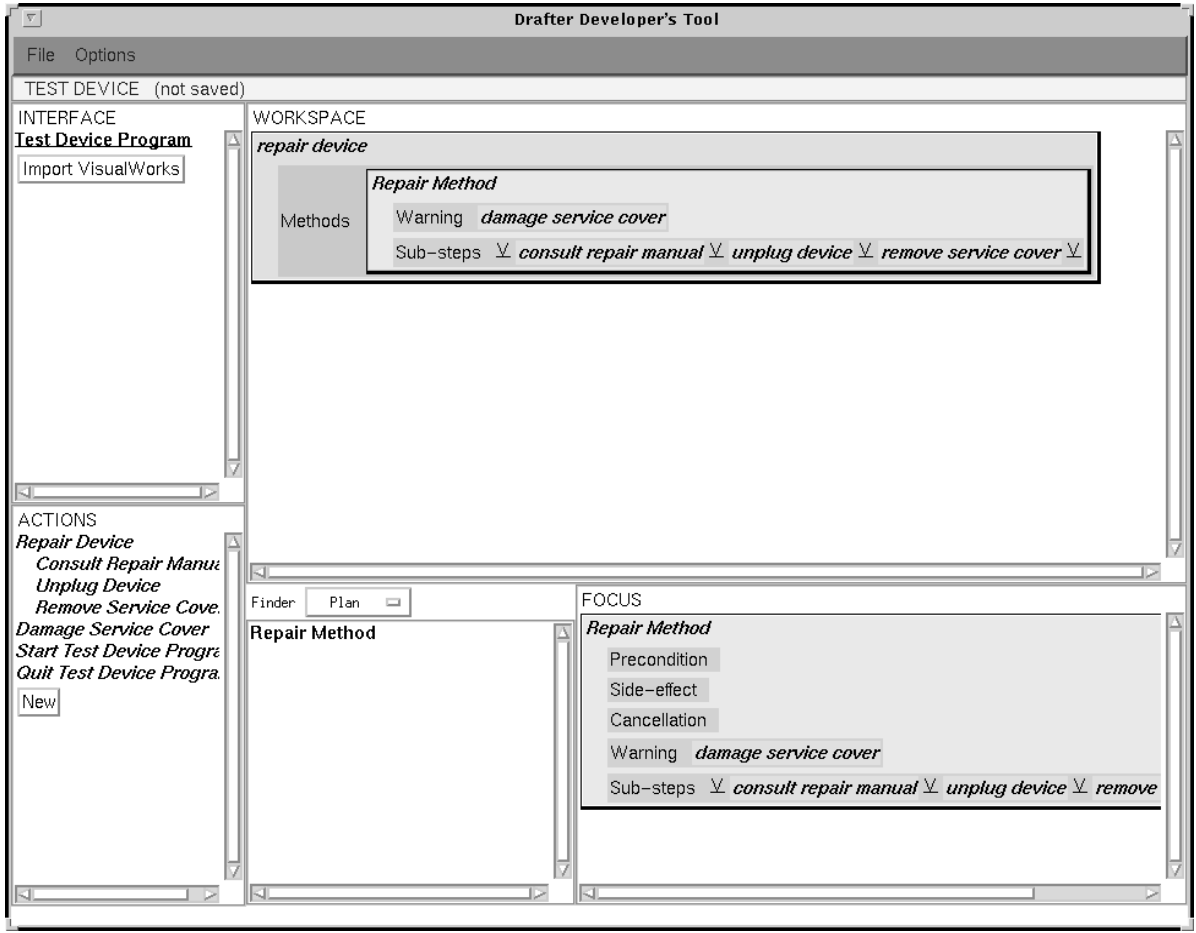,width=\textwidth}}
\caption{\drafter\ screen with the procedural structure for the example}
\label{example-graph}
\end{center}
\end{figure*}

In \drafter, technical authors specify the content of instructions in
a language independent manner using the \drafter\ specification tool.
This tool allows the authors to specify both the propositional
representations of the actions to be included, and the procedural
relationships between those propositions.  Figure~\ref{example-graph}
shows the \drafter\ interface after this has been done.  We will use
the procedure shown there as an example in this section, details on
how to build it can be found elsewhere (\cite{drafter-ieee96}).

The INTERFACE and ACTIONS panes on the left of
figure~\ref{example-graph} list all the objects and actions defined so
far.  These are all shown in terms of a pseudo-text which gives an
indication, albeit ungrammatical, of the nature of the action.  For
example, the main goal, ``repair device'', represents the action of
the reader repairing an arbitrary device.  This node may be expressed
in any number of different grammatical forms depending upon context.

The WORKSPACE pane shows the procedure, represented in an outline
format.  The main user goal of repairing the device is represented by
the largest, enclosing box.  Within this box, there is a single
method, called ``Repair Method'' which details how the repair should
be done.  There are three sub-actions: consulting the manual,
unplugging the device, and removing the cover. There is also a warning
slot filled with the action ``[reader] damage service cover''.  This
indicates that the reader should avoid damaging the service
cover.\footnote{Actually, this could also be interpreted as an {\em
ensurative\/} warning, meaning that the reader should make sure to
damage the service cover (although this is clearly nonsensical in this
case).  We have not yet analysed such expressions and thus do not
support them in \drafter.}

Neither the propositional nor the procedural information discussed so
far specify the three features needed by the decision network derived
in the previous section (i.e., intentionality, awareness, and safety).
At this point, we see no straight-forward way in which they could be
determined automatically (see Ansari's discussion of this issue
(\shortcite{ansari95:it})).  We, therefore, rely on the author to set
them manually.  \drafter\ allows authors to set generation parameters
on individual actions using a dialog box mechanism.
Figure~\ref{drafter-dialog-box} shows a case in which the author has
marked the following four features for the warning action ``damage
service cover'':

\begin{figure*}[t]
\begin{center}
\strut{\psfig{figure=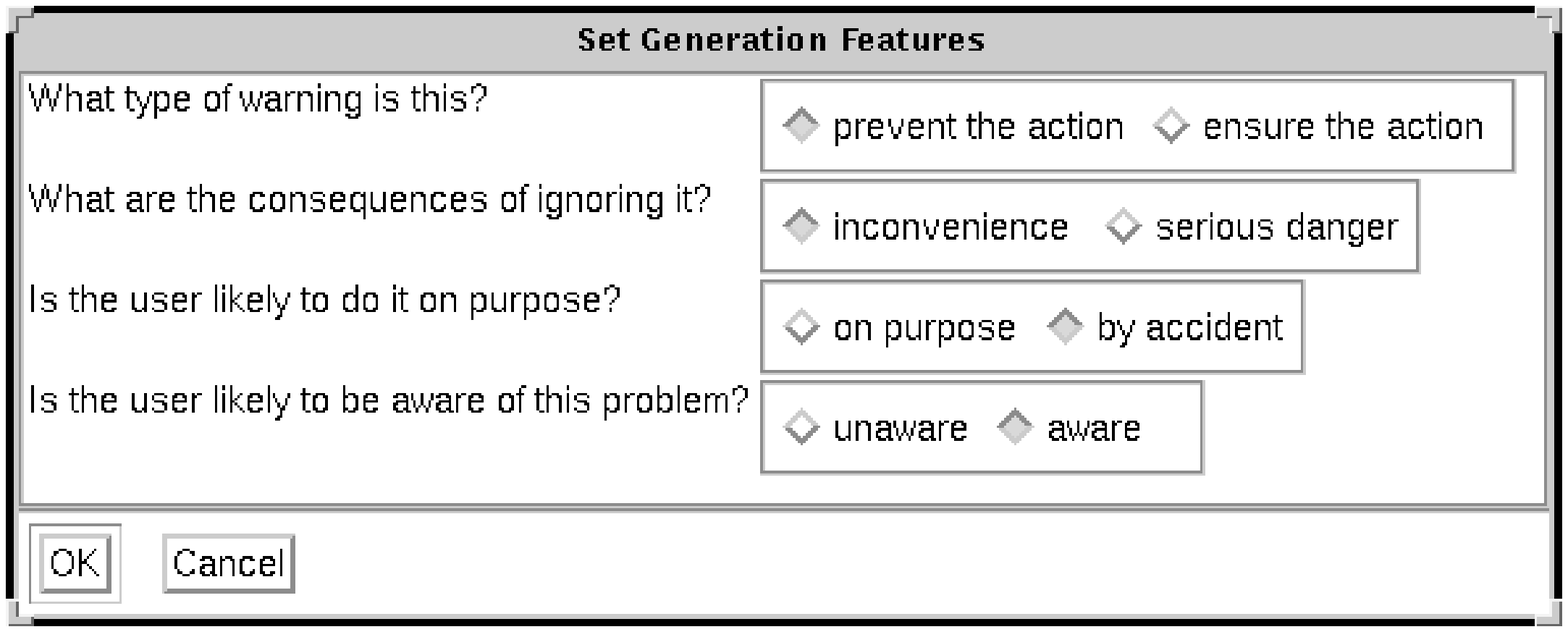,width=4.5in}}
\caption{The \drafter\ dialog box for setting the local parameters}
\label{drafter-dialog-box}
\end{center}
\end{figure*}

\begin{itemize}
\item The action is to be prevented, rather than ensured;
\item Performing the action would result in inconvenience, but not in
personal danger;
\item The user is likely to do the action accidentally, rather  than
consciously;
\item The user is likely to be aware that performing the 
action would create problems;
\end{itemize}

\subsection{Text Generation}

Once the input procedure is specified, the author may initiate text
generation from any node in the procedural hierarchy.  When the
technical author generates from the root goal node in
Figure~\ref{example-graph}, for example, the following texts are
produced:

\vbox{
{\em English:}

\begin{quote}
{\large\bf To repair the device} \\
1. Consult the repair manual. \\
2. Unplug the device. \\
3. Remove the service cover. \\
Take care not to damage the service cover. \\
\end{quote}
}

\vbox{
{\em French:}

\begin{quote}
{\large\bf R\'eparation du dispositif} \\
1. Se reporter au manuel de r\'eparation. \\
2. D\'ebrancher le dispositif. \\
3. Enlever le couvercle de service. \\
\'Eviter d'endommager le couvercle de service. \\
\end{quote}
}

\noindent

Note that the French version employs {\em \'eviter\/} ({\em avoid\/})
rather than the less common {\em prendre soin de ne pas\/} ({\em take
care not\/}).  This is possible because the French text is produced by
a separate micro-planning sub-network.  This sub-network was not based
on a corpus study of French preventatives, but rather was implemented
by taking the learned English decision tree, modifying it in
accordance with the intuitions of a French speaker, and automatically
constructing French systems from that modified decision tree.
Clearly, a corpus study French of preventatives is still needed, but
this does show \drafter's ability to make use of KPML's language
conditionalised resources.

Were we to replace the warning with other sorts of warnings, the
expression would also change according to the learned micro-planning
network.  If authors, for example, wish to prevent the reader from
performing the action of dismantling the frame of the device, and they
decide that the reader is unaware of this danger, that the action is
consciously performed and not unsafe, \drafter\ produces the following
text:

\vbox{
\begin{quote}
Do not dismantle the frame. \\
Ne pas d\'emonter l'armature. \\
\end{quote}
\vspace{-0.5cm}
}

If authors wish to prevent the reader from disconnecting the ground
connection, and they decide that the reader is unaware of this danger,
that the action would be unconsciously performed, and that the
consequences are indeed life-threatening,
\drafter\ produces the following text:

\vbox{
\begin{quote}
Never disconnect the ground. \\ Ne jamais d\'econnecter la borne de
terre. \\
\end{quote}
\vspace{-1cm}
}

\section{Conclusion}

In this paper we have discussed the use of machine learning techniques
for the automatic construction of micro-planning sub-networks.  We
demonstrated this for the case of preventative expressions in
instructional text.

We noted that because the automatic derivation of useful, well-defined
features for corpus analysis is beyond the current state of the art,
the painstaking process of corpus analysis must still be performed
manually.  As an example of how this can be done, we presented an
analysis of English preventative expressions.  We intend to continue
this part of the work by addressing more preventative forms,
addressing ensurative forms, and by extending the analysis to other
languages.

Although the analysis cannot be fully automated, we noted that the
derivation of decision networks from coded corpus examples can.  This
greatly simplifies the tasks of building and testing text planning
resources for new domains.  We intend to continue this part of the
work by applying the technique to larger portions of the planning
resources.

\subsection*{Acknowledgements}

The authors wish to acknowledge valuable discussions with Tony
Hartley, Xiaorong Huang, Adam Kilgarriff, C\'ecile Paris, Richard
Power, and Donia Scott, as well as detailed comments from the
anonymous reviewers.

\end{document}